\documentclass{article}

\usepackage[preprint]{neurips_2019}

\usepackage[utf8]{inputenc} 
\usepackage[T1]{fontenc}    
\usepackage{hyperref}       
\usepackage{url}            
\usepackage{booktabs}       
\usepackage{amsfonts}       
\usepackage{nicefrac}       
\usepackage{microtype}      
\usepackage{graphicx} 
\usepackage{caption}
\usepackage{subcaption}
\usepackage{enumitem}
\usepackage{amsmath}
\usepackage{titlesec}
\titleformat{\subsubsection}[runin]{\bfseries}{}{}{}[.]

\hypersetup{
    colorlinks=true,     
    urlcolor=cyan
}

\title{Spatiotemporal Action Recognition in Restaurant Videos}

\author{%
\\
  Akshat Gupta (akshatgu@andrew.cmu.edu)\\
  Milan Desai (mjdesai@andrew.cmu.edu)\\
  Wusheng Liang (wushengl@andrew.cmu.edu)\\
  Magesh Kannan (mkannan@andrew.cmu.edu) \\
  \textit{Carnegie Mellon University}
}

\begin{document}

\maketitle

\begin{abstract}
    Spatiotemporal action recognition is the task of locating and classifying actions in videos. Our project applies this task to analyzing video footage of restaurant workers preparing food, for which potential applications include automated checkout and inventory management. Such videos are quite different from the standardized datasets that researchers are used to, as they involve small objects, rapid actions, and notoriously unbalanced data classes. We explore two approaches – one involving the familiar object detector “You Only Look Once” (YOLO), and another applying a recently proposed analogue for action recognition, “You Only Watch Once” (YOWO). In the first, we design and implement a novel, recurrent modification of YOLO using convolutional LSTMs and explore the various subtleties in the training of such a network. In the second, we study the ability of YOWO’s three-dimensional convolutions to capture the spatiotemporal features of our unique dataset, which was generously lent by CMU-based startup \href{http://www.agot.ai}{Agot}.
\end{abstract}

\section{Introduction}

Advances in the field of computer vision have affected various industries, including healthcare, automotive, banking and many more. Availability of large data sets like ImageNet [9] has been a cornerstone of many of these advances. The most basic unit in an image from the perspective of computer vision is an {\it object}. The most basic task one can do then is to find if the object is present in the image or not. Thus, success in the task of object classification was been essential in further advancements. Many important architectures followed which helped in improving the performance of objection classification, including architectures like AlexNet [10], ResNet [11], MobileNet [12], etc .

The next obvious problem after finding different objects in the image is to detect the location of the object in the image, thus going towards the task of object detection. R-CNN [13] was one of the early successful architectures for object detection. It was a two-stage network, where the network first performed object classification and then finalized the bounding boxes around the image. This meant training two separate networks for doing the task of object detection. The need for a two-stage network was eliminated by YOLO [2], which was the first end-to-end object detection network. It was surprisingly fast and produced competitive results in real-time.

A lot of work in computer vision has also been done with videos. A video is a time series of images, and thus has a spatial and a temporal component. The most basic unit that we are concerned with for videos, from a computer vision standpoint, is a spatial object that also has a time component. Such a unit is what makes up an {\it action}. The task of action classification is one of the most basic tasks when dealing with videos. Analogous to the work done with images, people have worked in action localization, scene understanding, video summarization, and more.

The major data sets available for action classification are UCF [14], HMDB [15] and Kinetics dataset [16]. Before the Kinetics dataset [16], the available datasets were not large enough to train 3D CNN's, which have way more parameters than their 2D counter parts. After the introduction of the kinetics dataset, we were finally able to train deep 3D convolutional networks [17], similar in size and performance to the architectures trained on ImageNet. Thus, action classification has achieved similar success as compared to the object classification and detection. It remains to be seen that this will analogously correspond to success in tasks like action localization, video summarization, etc.

Our work lies in the domain of action localization, where we classify spatio-temporal actions as well as localize them on a dataset of restaurant videos. Here, action recognition has the potential to automate tasks like checkout, inventory management, and quality insurance. A recent Carnegie Mellon University - based startup, Agot, aims to specialize in just that, and provided us a sample of annotated footage of workers preparing food at a fast, carry-out style Sushi restaurant. The data is unlike the previously mentioned standardized datasets; they have fast-moving actions, small bounding boxes, poor class balance, and imperfect bounding boxes and labels.

Our project explores multiple approaches to applying action recognition to a production application using a nonstandardized dataset. We design a recurrent version of the popular object detection network, YOLO, referred from here on as {\it Recurrent YOLO}. We also work with YOWO (You only Watch Once) [1], which is the first end-to-end architecture for spatio-temporal action localization, and modify it for our task.

\section{Related Work}
\subsection{You Only Look Once}
YOLO ("You Only Look Once") is a state-of-the-art object detection framework first published in 2015 by Joseph Redmon et al [2]. It is an end-to-end solution comprising a single network; an image frame goes in, and bounding box and class predictions come out. Since the initial publication, substantial improvements have been made to the framework, culminating in YOLOv3, published in 2018 [4]. In addition to achieving a high classification accuracy, YOLO is fast, allowing real-time object detection in videos. The relatively simple pipeline consists of a single convolutional neural network, detecting objects in an image in a single pass.

The basic architecture of YOLO is a CNN with 24 convolutional layers and two fully connected layers. The network divides the input image into an $S \times S$ grid, where each grid cell predicts $B$ bounding boxes and a single class. For an object that spans multiple grid cells, the cell containing the object's center is the responsible predictor. Therefore, the output of the network has shape $S \times S \times (5B + C)$, encapsulating the bounding box coordinates, confidence score and class probabilities for each grid cell. Boxes with a high enough confidence score (also known as "objectness") are considered to have detected objects.

During training, intersection-over-union (IOU) with the ground truth is used to select only one box in the cell as the object's predictor. A specialized loss function combines classification loss, localization loss, and confidence loss, which measure error in the class predictions, box coordinates, and confidence/objectness scores, respectively. During detection, box predictions with confidence scores below a configured threshold are filtered out, and non-maximum suppression is use to further reduce the number of detected objects [2].

\subsection{YOLOv2}
While YOLO has a high classification accuracy, it suffers from localization errors and low recall compared to other object detection systems [3]. YOLOv2 improves upon its predecessor by introducing batch normalization, a better training procedure, and most importantly, anchor boxes. Anchor boxes stem from the observation that most objects of the same category have similar sizes and shapes. Rather than predicting the box shape directly, YOLOv2 computes offsets of predetermined boxes, termed anchor boxes, or 'priors'. A fixed number of anchor boxes are precomputed using K-means clustering on the training set.

\subsection{YOLOv3}
The most recent iteration of YOLO further improves performance by using a 53-layer residual network as the feature extractor. It also outputs predictions at three different scales to better support the detection of objects of varying sizes. At each scale, the feature extraction results are passed through a 1x1 convolution and a 'detection' layer, which computes bounding boxes based on the configured anchor boxes. Each of the three detection layers is configured with three anchor boxes, so the network has nine anchor boxes in total [4].

\subsection{You Only Watch Once}

YOWO ("You Only Watch Once") [1] is a unified CNN architecture for real-time video spatiotemporal action localization proposed by Okan Köpüklü, et al. in December 2019 and is still under review for publication. Instead of separating the task into 2 stages, proposal stage and classification and localization stage like Faster R-CNN [5], which has also inspired most state-of-art works on these tasks, YOWO is an end to end network. It has two branches. One branch is used for extracting spatial features from the key frame, another is for spatiotemporal features for the whole clip. It is the only architecture trained end to end for spatio-temporal action as far as literature review tells us.

The YOWO network is both fast (62 fps on 8-frame clips) and the best performing network for action localization till date, outperforming the previous state-of-art results on J-HMDB-21 and UCF101-24. The architecture is also flexible in having different backbones for the 2D feature extractor and 3D feature extractor backbones depending on our application. We've experimented with a number of 3D CNN backbones in our work. 
\subsection{YOWO Architecture}

The architecture of YOWO network can be divided into three parts: feature extractor, CFAM (channel fusion and attention mechanism), and bounding box regression. The feature extractor in has two branches for feature extraction. It uses a 2D CNN for key frame spatial features and a 3D CNN for spatiotemporal features, following which the 2D and 3D features are aggregated together. The 2D feature extractor utilizes Darknet-19 (also used in YOLO) as its backbone for balance between accuracy and efficiency. The 3D feature extractor uses 3D-ResNext-101 as its basic architecture due to its high performance when tested on multiple datasets [6].

The CFAM (Channel Fusion and Attention Mechanism) module is a channel fusion and attention mechanism for integrating channel attention with the feature structure. Before outputting from the CFAM block, YOWO applies another two convolutional layers to help mix the features. This mechanism can allocate more weights (attention) to correlated channels for a given channel. After getting the fused and attention-weighted feature map, YOWO follows the bounding box regression method for YOLOv2 [3].

\section{Dataset}

\begin{figure}
    \centering
    \begin{subfigure}[b]{0.45\textwidth}
        \centering
        \includegraphics[width=\textwidth]{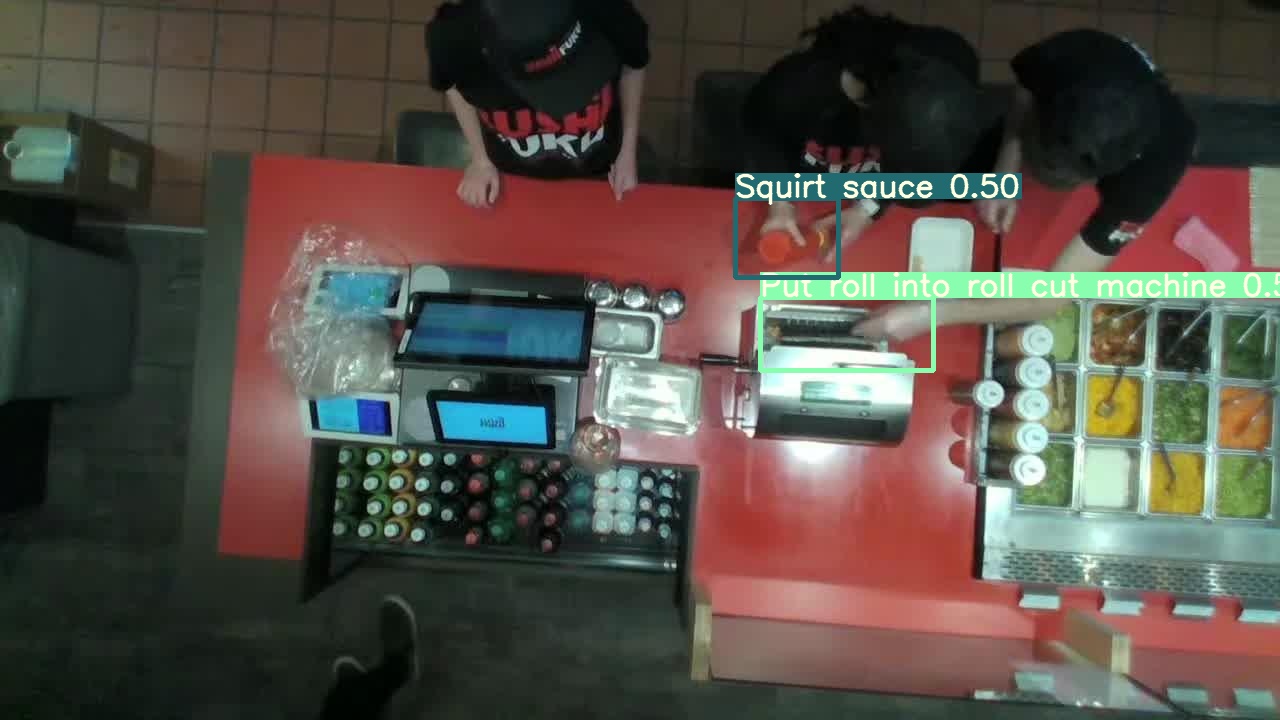}
    \end{subfigure}
    \hfill
    \begin{subfigure}[b]{0.45\textwidth}
        \centering
        \includegraphics[width=\textwidth]{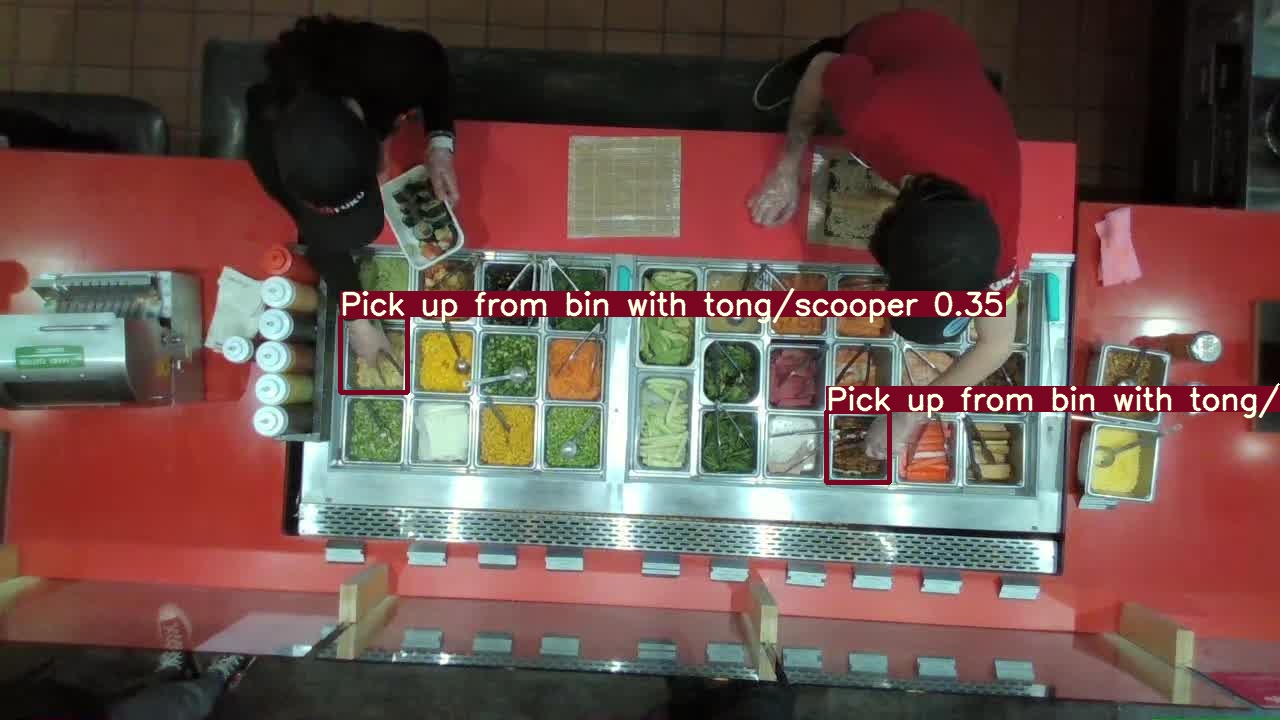}
    \end{subfigure}
    \caption{Example predictions from each vantage point. These particular boxes and labels were predicted by the baseline YOLO object detector.}
    \label{fig:example_data}
\end{figure}

\begin{figure}
\includegraphics[scale=0.8]{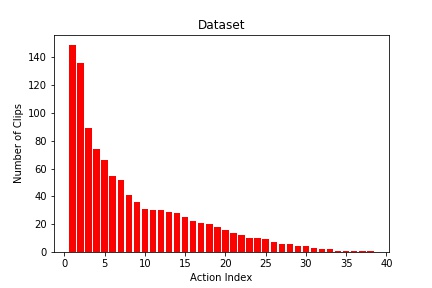}
\caption{Distribution of actions in the dataset. The action index assigns a number to each action. The index and the corresponding action for the top 5 actions (with the 5 most number of clips) are as follows: 'pick up from bin with tong or scooper', 'put item into meal using tongs', 'put tongs or scooper back in bin', 'operating POS', 'put item into meal using hands'.}
\label{fig:dataset}
\end{figure}

Our training data, provided by Agot, consists of labeled video footage of the same restaurant from two different vantage points (see figure \ref{fig:example_data}). The combined 100 minutes of 720p footage has close to 36,000 still image frames, of which roughly half have no labeled bounding boxes. The dataset was further split into clips of varying length, where the space-time coordinates of the actions and labels were given for each clip. Out of the total set of 45 actions, we had labeled data for 38 actions, of which 24 have 10 or more labeled clips. The dataset is highly imbalanced, with some actions having far fewer examples than others. Figure \ref{fig:dataset} shows the number of clips per action.

\section{Approach}
\subsection{Recurrent YOLO}

\begin{figure}
\includegraphics[width=0.55\linewidth]{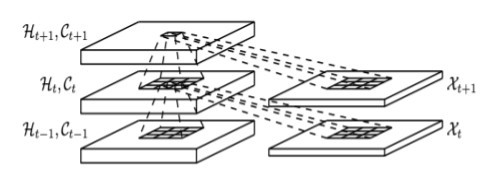}
\caption{Convolutional LSTM block. All weight multiplications of a traditional LSTM are replaced with convolution operators. Figure borrowed from [19].}
\label{fig:convlstm}
\end{figure}

\begin{figure}
\includegraphics[width=0.85\linewidth]{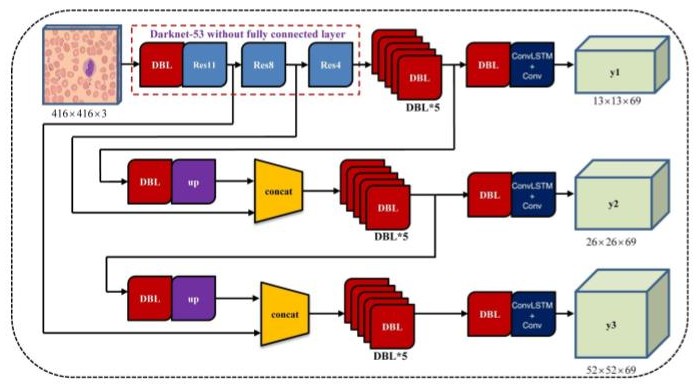}
\caption{Recurrent YOLO architecture. We insert ConvLSTM layers prior to final convolution of each of the three detection layers (which correspond to the three scales of prediction in YOLOv3).}
\label{fig:recurrentyolo}
\end{figure}

As YOLO is a state-of-the-art object detector capable finding spatial features in images, it is conceivable that a recurrent modification to the architecture could find spatiotemporal features. We explored this approach by designing a network that adds long short-term memory (LSTM) [18] after the spatial feature detectors and prior to the output layers of YOLOv3. To ensure that the LSTM layer could remember spatial features over time, we implemented a convolutional LSTM, or \textbf{ConvLSTM} [19]. A ConvLSTM replaces the weight multiplications of traditional LSTMs with convolutional operators, as shown in figure \ref{fig:convlstm}. In our implementation, we additionally replaced the peephole Hadamard products with convolutions to avoid dependence on the image dimensions. The modified state transitions become as follows, where $*$ represents a convolution and $\circ$ represents the Hadamard product:

\begin{gather}
\begin{aligned}
    i_t &= \sigma(W_{xi} * X_t + W_{hi} * H_{t-1} + W_{ci} * C_{t-1} + b_i)\\
    f_t &= \sigma(W_{xf} * X_t + W_{hf} * H_{t-1} + W_{cf} * C_{t-1} + b_f)\\
    C_t &= f_t \circ C_{t-1} + i_t \circ \text{tanh}(W_{xc} * X_t + W_{hc} * H_{t-1} + b_c)\\
    o_t &= \sigma(W_{xo} * X_t + W_{ho} * H_{t-1} + W_{co} * C_{t} + b_o)\\
    H_t &= o_t \circ \text{tanh}(C_t)
\end{aligned}
\end{gather}

Using a ConvLSTM also avoids an explosion in the number of parameters learned by the model. With a traditional LSTM, we would be forced to flatten the outputs of the convolutional layers and have weights mapping each input pixel to each output pixel. The number of parameters required for an LSTM with $m \times n$ input and hidden size $h$ is proportional to $(h m n)^2$, but only $(h k)^2$ for a ConvLSTM with kernel size $k$.

The recurrent architecture is shown in figure \ref{fig:recurrentyolo}. Notice that the ConvLSTMs are inserted before the final convolution of each YOLO output. They take as input the spatial features learned from the previous layers and try to build a spatiotemporal representation prior to generating the final boxes. Our ConvLSTM was implemented from scratch in PyTorch and supports dropout, layer stacking, and bidirectionality.

\subsection{Implementation}
After many attempts at finding an implementation suitable to our proposed revisions, we used a PyTorch implementation of YOLOv3 by Ultralytics LLC as our baseline. The code, including our ConvLSTM implementation and other modifications, may be found \href{https://github.com/milandesai/yolov3}{here}.

\subsection{Training}
To train the network, we first used k-means clustering to find the anchor box dimensions that best match our dataset. We then pretrained a nonrecurrent YOLOv3 network on our video frames and ported the weights of the convolutional layers to the new model. The input to the recurrent network at each iteration step was a sequence of image frames. Layers prior to and after the LSTMs treated the entire sequence as a single batch of frames, while the recurrent layers treated it as a single sequence with batch size of 1. We used the same loss function as in YOLO, and computed this loss by comparing the predictions and targets for each frame independently.

\subsection{Hybrid Shuffling}
For the recurrent layer to effectively find temporal features, the input must be a sequentially ordered list of frames. A naive solution would be to pass the training data in without shuffling, but this reduces the generalizability of the model and the effectiveness of stochastic gradient descent. Another problem is that nearly half the frames in the training set do not correspond to any labeled actions, which could substantially slow down learning. Instead, we designed a hybrid shuffling solution whereby we first preprocessed the dataset into clips, and shuffled the set of clips but not the frames themselves.

The procedure is as follows. We first segmented the ordered list of frames into clips of continuous action sequences, padded with 30 frames before the start and 30 frames after the end of the sequence. Aside from this padding and small breaks within a sequence, no frames without labeled actions were loaded. Clips were allowed to overlap, and their lengths ranged from 100 to 1000 frames. We loaded the clips such that their frame order was preserved, but allowed the clips themselves to be shuffled. They were too large to be fed into the network as entire batches, so they were input in smaller subsequences at a time, with the LSTM state preserved and forwarded until the end of the clip.

This training procedure allowed the convolutional LSTM layers to remember features for the duration of entire action sequences and reduced the size of the training set by 40\%. However, it also precluded the use of bidirectional LSTMs and complicated the use of multiple of GPUs. Allowing bidirectionality would not allow the LSTM state to be forwarded between subsequences of a clip without using an extraordinary amount of memory, since the hidden states of each frame for the entire duration of a clip would have to be saved. Using multiple GPUs requires enforcing that all frames of a given clip are loaded on the same GPU, whose LSTM's hidden state is independent of that of others.

\subsection{Validation}
For pretraining nonrecurrent YOLO, we randomly selected 20\% of the frames to be in our validation set, but to avoid having the network simply learning to reconstruct isolated missing frames, we ensured that every 120 frames of footage (4 seconds in duration) was kept in the same dataset, training or test. For training recurrent YOLO, we randomly chose 20\% of the preprocessed clips to be in the validation set. We did not have a separate validation and test set due to the dearth of training data.

\subsection{YOWO}

To test how our data performs on a state-of-art model and also provide a baseline for our novel recurrent YOLO network, we ran \href{https://github.com/wei-tim/YOWO}{YOWO}, proposed and implemented in [1] with our data. Instead of combing a region proposal network and a classification network to do the spatiotemporal localization task, YOWO tried to directly stack and merge the 2D spatial features and 3D spatiotemporal features with a channel fusion and attention mechanism. The 2D features are extracted with Darknet 19, 3D features can be extracted with ResNext101, ResNet50, ShuffleNet etc. 

To train with YOWO, we first separated our long videos and annotations to short video clips and corresponding label files, then they are put into different folders by their action labels. We applied the YOWO pretrained weights on UCF101-24 dataset to the network and tested with both freezing the 2D and 3D backbone weights frozen or fine tuning the weights. Our data is loaded as a short video clip of 16 frames duration, the video clip is fed directly into 3D CNN backbone, the last frame of the clip is the key frame, which is fed into the 2D CNN. Data augmentation such as cropping and shifting is applied. 

When predicting the bounding boxes, YOWO takes advantage of YOLOv2’s [3] strategy. In the last layer, for each grid cell, 5 boxes will be predicted, each box will provide box size and coordinates, as well as a confidence score, and probability for each action class. When predicting the box size, 5 anchor box priors are given, YOWO will first pick out the most possible anchor box size by comparing with the ground truth, then the bounding box size will be predicted based on this anchor box, which makes anchor boxes an essential parameter to customize. We used K-means clustering to pick the anchor box priors for our dataset. The loss function is the same as YOLO, different contribution scales are given to different components of a joint loss, the whole network can be optimized end-to-end.

\section{Results}

We evaluated the results of our experiments on YOWO and recurrent YOLO by measuring precision, recall, and mean average precision (mAP) at an intersection-over-union (IOU) threshold of 0.5 on the validation set. Noting the large imbalance in the class distribution of our dataset, we report these metrics only for the top 5 and top 24 most frequent actions.

The control for our experiments was the training of a standard, nonrecurrent YOLOv3 model on our dataset. Training took 15 minutes per epoch using four Nvidia Tesla V100 GPUs and plateaued after 30 epochs. We obtained a top-5 precision of 54.2\%, recall of 66.9\%, and mAP of 59.7, and a top-24 precision of 59.3\%, recall of 68.6\%, and mAP of 64.3. This baseline model achieved highest mAP on the actions, "Roll a sushi burrito", "Using a cellphone", and "Operating a Sushi Roll Cutter", at over 97\%. 

Recurrent YOLO trained substantially slower, at 2 hours per epoch on a single Tesla V100, twice the per-GPU time of nonrecurrent YOLO. It also failed to converge and produce meaningful output, despite various techniques including adding additional stacked LSTM layers, freezing the pretrained weights, and modifying hyperparameters.

We trained YOWO with many different backbones. Training YOWO with top-5 classes for one epoch with different 3D backbones took between 40 and 90 minutes on a single NVIDIA Tesla T4 GPU. Training the top-24 classes took 4 hours on the same GPU, and 2 hours with 2 GPUs. The plateau was reached within 5 epochs. 

Table 1 compares the results of YOWO and baseline YOLO for the top 5 and top 24 classes. For the top 5 classes, we used Darknet-19 as the 2D CNN backbone. We worked with several different 3D backbones including the ResNet18, ResNet50, ResNext101 and MobileNet. The results are shown in Table 2. We got the highest localization recall of 61.1\% and classification accuracy of 91.3\% with ResNet50 as the 3D backbone. An important thing we noticed was that once a bounding box was localized, the classification accuracy into an action is quite high in all the cases. We observe the same thing when we run YOWO for all the 24 classes. All the results in Table 2 use custom anchor boxes learned through k-means clustering except the last case. 

For top 24 classes, we tried combinations of customizing anchor boxes and freezing 2D and 3D backbone pre-trained weights. We got the highest localization recall of 58.33\%, classification accuracy of 94.37\% and mAP of 29.47. Table 3 shows the results of the different customizations we tried. We find that using custom anchor boxes always helped our results. Using pre-trained YOWO weights produced the best results when used with our custom anchor boxes.

\begin{center}
    \includegraphics[width=\textwidth]{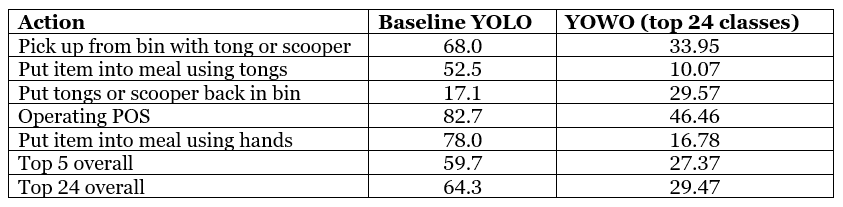}
\end{center}

\small{\textbf{Table 1. mAP Comparison between Baseline YOLO and YOWO.} The first 5 rows for the top-5 individual actions, followed by top-5 actions, and top-24 actions.}

\begin{center}
    \includegraphics[width=\textwidth]{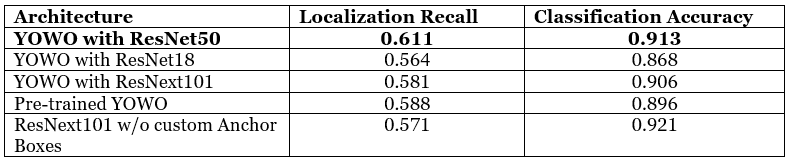}
\end{center}

\small{\textbf{Table 2. Top 5 Classes Performance on YOWO.} Highest recall was achieved with ResNet50 as 3D backbone. Darknet 19 is used as 2D backbone for all of them. Pretrained YOWO weights were trained with UCF101-24. Custom anchor boxes were obtained with K-means clustering.}

\begin{center}
    \includegraphics[width=\textwidth]{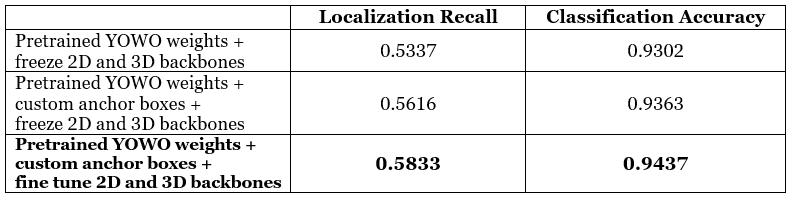}
\end{center}

\small{\textbf{Table 3. Top 24 Classes Performance on YOWO.} Highest recall and classification accuracy were achieved with custom anchor boxes and fine tuning the 2D and 3D backbones. Pretrained YOWO weights were trained with UCF101-24 dataset, Darket 19 is used as 2D backbone.}

\normalsize

\section{Analysis}

At a first glance, the baseline results from nonrecurrent YOLO are surprisingly good for a pictorial object detector on a video action dataset. Upon inspection, the results are not surprising at all. Simply, the object detector has learned to detect objects - cell phones, sushi rolls, machines, and hands (based on the location of the hand, it can detect the operation of a specific machine, for example). Naturally, actions that have a more temporal aspect, such as picking items up and putting them down, fared worse. However, the results show that YOLO can construct spatial features that are useful for action recognition, lending support to the notion of a recurrent YOLO.

Our custom convolutional-LSTM based YOLO architecture did not fare as well, as we were unable to extract any meaningful improvements or results during training. There are several possible reasons for this. One is that the network simply needs more time to learn, as the LSTMs add substantial complexity to the model. Giving the network additional training time, however, was hindered by the fact that not only did the LSTMs double the per-GPU epoch time for one hour to two hours, but also the training procedure prevented us from using multiple GPUs. Effectively, this represented an 8x slowdown compared to the baseline experiment and prevented a high-epoch investigation, even after our use of a training procedure that cut the dataset in nearly half by preprocessing it into clips.

However, multiple GPUs could be supported by carefully implementing a distribution mechanism that sends subsequences of the same clip to the same GPU, in order, and resets the LSTM hidden and memory states on a particular GPU only after a new clip is loaded. Our project mentor has also suggested that we use a sparse frame representation as input to the network instead of every frame in a sequence [20]. Any given frame in the ordered dataset will be very similar to its close neighbors, so we could substantially reduce the dataset size by discarding $N-1$ out of every $N$ frames.

Another reason for the failure of recurrent YOLO could be that the LSTMs are inserted too late in the architecture, at nearly just prior to the detection layers. The motivation behind this design was to use YOLO weights initialized from the pretrained model to capture the most complete spatial feature representations, which are known to work well for detection, and use the recurrence to find the temporal relationships between those features. However, it may be that the compression is too great by the time the input reaches the LSTMs for them to extract the relevant temporal features. Moving the recurrent layer to earlier in the network merits further investigation.

The performance of YOWO on our dataset is not as good as it achieved on UCF101-24 or J-HMDB-21, obtaining high classification accuracy but suffering from low precision and recall. This indicates the model has trouble finding actions, but when it does, it knows how to classify them. The imbalance of data is a major reason for this performance. Furthermore, from observation, comparing with YOWO anchor boxes, our box sizes lack variation in size; most of them are of similar sizes and relatively small. On the one hand, these boxes picked by k-means clustering represent the most common box sizes for our data, but on the other hand, the lack of divergence in box size might cause extra hardship when doing the prediction. As YOWO will first pick the most similar anchor box prior for each ground truth box, when most of the anchor boxes are of similar size, some of them might have not been used at all, while some larger boxes become harder to predict precisely.

Another possible reason that our performance cannot reach that of UCF101-24 is that because our videos are taken in only 2 scenes, most of the video clips have similar or the same environments. We should not expect the model to learn from the surrounding instead of the action itself, but the lack of background information might be a reason of reduced performance on identification.

\section{Conclusions}
Our goal was to experiment with multiple approaches for spatiotemporal action recognition on a dataset of restaurant videos. We found that baseline YOLO, which does not construct temporal features, can still find spatial features that are useful for action detection. We made a recurrent modification using convolutional LSTMs with a specialized training procedure but were unable to successfully train it. We also found that YOWO can learn the spatiotemporal features from our videos and achieve good results on some classes but could not reach the overall performance it achieved on other datasets. The imbalance of data and similarity of video environments may be partially accountable. Overall, these results demonstrate that applying spatiotemporal action recognition to a non-standard dataset for a production application is a nontrivial task. Future research may expand on the techniques described here to improve performance. Continued exploration for a recurrent YOLO could involve migrating the recurrent layers to earlier in the network and implementing techniques to improve training speed. As for YOWO, trying with additional different backbones, changing the strategy of picking anchor box priors, and using learnable thresholds may improve performance. 

\section*{References}
[1] Köpüklü, O., Wei, X.,\ \& Rigoll, G.\ (2019) You Only Watch Once: A Unified CNN 
Architecture for Real-Time Spatiotemporal Action Localization. 
{\it arXiv preprint arXiv:1911.06644}.\label{YOWO}

[2] Redmon, J., Divvala, S., Girshick, R.\ \& Farhadi, A.\ (2016) You Only Look Once: Unified, Real-Time Object Detection. {\it Proceedings of the IEEE conference on computer vision and pattern recognition}, pp.\ 779--788.

[3] Redmon, Joseph, and Ali Farhadi.
YOLO9000: better, faster, stronger.
{\it Proceedings of the IEEE conference on computer vision and pattern recognition}. 2017.

[4] Redmon, J.\ \& Farhadi, A.\ (2018) YOLOv3: An incremental improvement. 
{\it arXiv preprint arXiv:1804.02767}.

[5] Shaoqing Ren, Kaiming He, Ross Girshick, and Jian Sun.
Faster R-CNN: Towards Real-Time Object Detection with Region Proposal Networks.
{\it Proceedings of the IEEE conference on computer vision and pattern recognition}.2017.

[6] Hara, Kensho, Hirokatsu Kataoka, and Yutaka Satoh.
Can spatiotemporal 3d cnns retrace the history of 2d cnns and imagenet?.
{\it Proceedings of the IEEE conference on Computer Vision and Pattern Recognition}. 2018.

[7] Xie, Saining, et al.
Aggregated residual transformations for deep neural networks.
{\it Proceedings of the IEEE conference on computer vision and pattern recognition}. 2017.

[8] Lin, Tsung-Yi, et al.
Focal loss for dense object detection.
{\it Proceedings of the IEEE international conference on computer vision}. 2017.

[9] Deng, Jia, et al. "Imagenet: A large-scale hierarchical image database." 2009 IEEE conference on computer vision and pattern recognition. Ieee, 2009. \label{ImageNet}

[10] Krizhevsky, Alex, Ilya Sutskever, and Geoffrey E. Hinton. "Imagenet classification with deep convolutional neural networks." Advances in neural information processing systems. 2012. \label{AlexNet}

[11] He, Kaiming, et al. "Deep residual learning for image recognition." Proceedings of the IEEE conference on computer vision and pattern recognition. 2016. \label{ResNet}

[12] Howard, Andrew G., et al. "Mobilenets: Efficient convolutional neural networks for mobile vision applications." arXiv preprint arXiv:1704.04861 (2017). \label{MobileNet}

[13] Girshick, Ross, et al. "Rich feature hierarchies for accurate object detection and semantic segmentation." Proceedings of the IEEE conference on computer vision and pattern recognition. 2014.

[14] Soomro, Khurram, Amir Roshan Zamir, and Mubarak Shah. "UCF101: A dataset of 101 human actions classes from videos in the wild." arXiv preprint arXiv:1212.0402 (2012).

[15] Kuehne, Hildegard, et al. "HMDB: a large video database for human motion recognition." 2011 International Conference on Computer Vision. IEEE, 2011.

[16] Kay, Will, et al. "The kinetics human action video dataset." arXiv preprint arXiv:1705.06950 (2017).

[17] Hara, Kensho, Hirokatsu Kataoka, and Yutaka Satoh. "Can spatiotemporal 3d cnns retrace the history of 2d cnns and imagenet?." Proceedings of the IEEE conference on Computer Vision and Pattern Recognition. 2018.

[18] Hochreiter, Sepp, and Jürgen Schmidhuber. "Long short-term memory." Neural computation 9.8 (1997): 1735-1780.

[19] Xingjian, S. H. I., et al. "Convolutional LSTM network: A machine learning approach for precipitation nowcasting." Advances in neural information processing systems. 2015.

[20] Leo Chenghui Li (project mentor) in discussion with the authors, May 2020.

\end{document}